\documentclass[11pt,a4paper]{article}
\usepackage[hyperref]{acl2021}
\usepackage{times}
\usepackage{latexsym}
\usepackage{algorithm, algorithmic}
\usepackage{amsmath} 
\usepackage{amssymb}
\usepackage{multirow}
\usepackage{makecell}
\usepackage{hhline}
\usepackage{xspace}
\usepackage{booktabs}
\usepackage{float}
\usepackage{graphicx}

\usepackage{amssymb}
\usepackage{pifont}
\usepackage{bm}
\usepackage{enumitem}
\usepackage{diagbox}
\usepackage{enumitem}
\setlist[itemize]{leftmargin=*}
\setitemize[1]{itemsep=0pt,partopsep=0pt,parsep=\parskip,topsep=0pt}
\usepackage{microtype}

\aclfinalcopy 
\def\aclpaperid{4080} 

\title{Word Graph Guided Summarization for Radiology Findings}

\author{%
Jinpeng Hu$^{\spadesuit\heartsuit}$, \hspace{0.2cm}
Jianling Li$^{\diamondsuit}$, \hspace{0.2cm}
Zhihong Chen$^{\spadesuit\heartsuit}$, \hspace{0.2cm}
Yaling Shen$^{\spadesuit}$, \\
\textbf{Yan Song}$^{\spadesuit\heartsuit}$, \hspace{0.2cm} \textbf{Xiang Wan}$^{\heartsuit}$, \hspace{0.2cm} \textbf{Tsung-Hui Chang}$^{\spadesuit\heartsuit}$ \\
$^{\spadesuit}$The Chinese University of Hong Kong (Shenzhen),\\ 
$^{\heartsuit}$Shenzhen Research Institute of Big Data, \hspace{0.2cm}
$^{\diamondsuit}$National University of Defense Technology \\
\texttt{
$^\spadesuit$\{jinpenghu, zhihongchen, yalingshen\}@link.cuhk.edu.cn} \\
\texttt{$^\diamondsuit$jianlingl@nudt.edu.cn} \hspace{0.2cm}
\texttt{$^{\heartsuit}$wanxiang@sribd.cn} \\
\texttt{$^{\spadesuit}$\{songyan,changtsunghui\}@cuhk.edu.cn}
}

\begin{document}
\maketitle
\renewcommand{\thefootnote}{\arabic{footnote}}
\begin{abstract}
Radiology reports play a critical role in communicating medical findings to physicians. In each report, the impression section summarizes essential radiology findings.
In clinical practice, writing impression is highly demanded yet time-consuming and prone to errors for radiologists. Therefore, automatic impression generation has emerged as an attractive research direction to facilitate such clinical practice.
Existing studies mainly focused on introducing salient word information to the general text summarization framework to guide the selection of the key content in radiology findings.
However, for this task, a model needs not only capture the important words in findings but also accurately describe their relations so as to generate high-quality impressions.
In this paper, we propose a novel method for automatic impression generation, where a word graph is constructed from the findings to record the critical words and their relations, then a \underline{W}ord \underline{G}raph guided \underline{Sum}marization model (\textbf{\textsc{WGSum}}) is designed to generate impressions with the help of the word graph.
Experimental results on two datasets, \textsc{OpenI} and \textsc{MIMIC-CXR}, confirm the validity and effectiveness of our proposed approach, where the state-of-the-art results are achieved on both datasets.
Further experiments are also conducted to analyze the impact of different graph designs to the performance of our method.\footnote{Our code and the best performing models are released at \url{https://github.com/jinpeng01/WGSum}.}
\end{abstract}

\section{Introduction}
\begin{figure}[t]
\centering
\includegraphics[width=0.48\textwidth, trim=0 20 0 8]{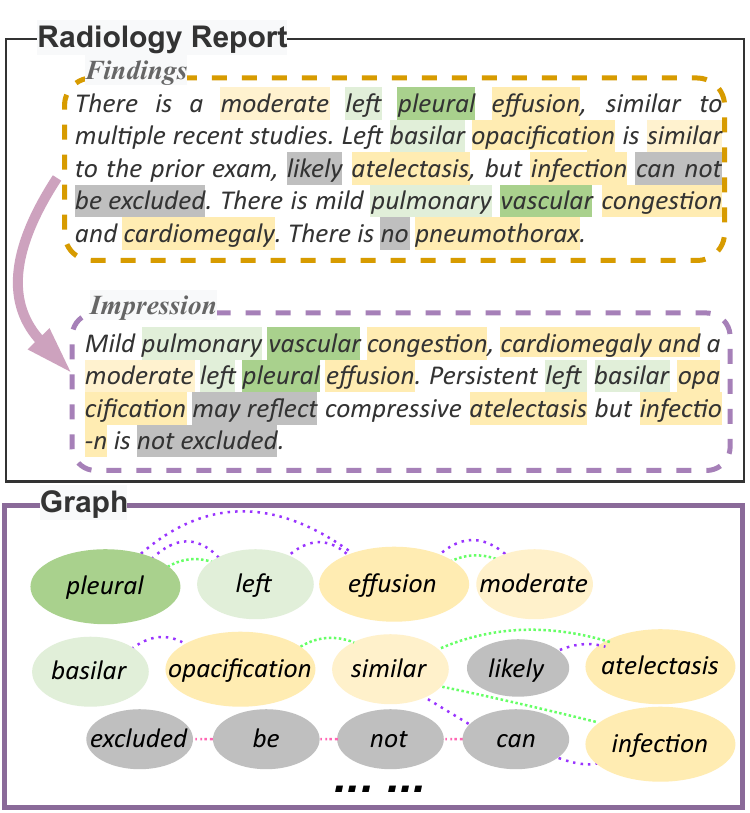}
\vspace{0.0001cm}
\caption{An example radiology report including its \textsc{Findings} and \textsc{Impression}, and a word graph constructed from the \textsc{Findings}. Different colored edges in the graph represent different types of relations.
The curved arrow indicates the AIG task to generate the \textsc{Impression} from the \textsc{Findings}.}
\label{fig:example}
 \vskip -1.5em
\end{figure}
\begin{figure*}[t]
\centering
\includegraphics[width=0.92\textwidth, trim=0 10 0 0]{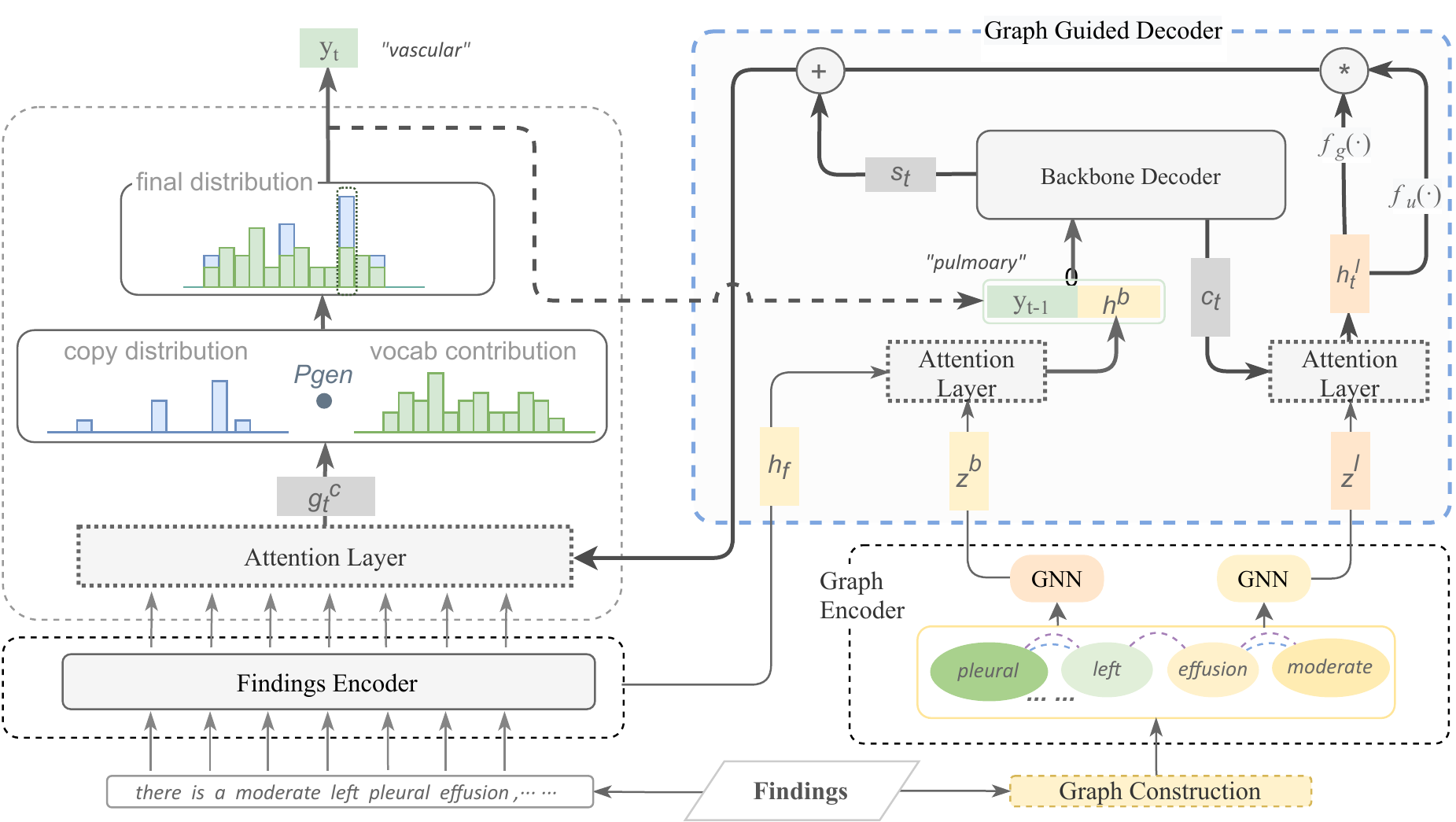}
\caption{The overall architecture of our proposed method with an example input and sample outputs at the $t-1$ and $t$ step. The \textsc{Findings} Encoder and the Graph Encoder are shown in different grey dashed boxes with their details omitted. The Graph Guided Decoder is illustrated in the blue dashed box.}
\label{fig:architecture}
 \vskip -1em
\end{figure*}

A radiology report usually contains a findings section (\textsc{Findings}) describing detailed medical observations and an impression section (\textsc{Impression}) summarizing the most critical observations.
In practice, the \textsc{Impression} is an essential part and plays an important role in delivering critical findings to clinicians.
Therefore, summarizing \textsc{Findings} helps to locate the most prominent observations so that the automatic process of doing so greatly eases the workload of radiologists. Recently, many methods are proposed for automatic impression generation (AIG) \cite{iemrr, zhang2018learning, attend}, which are mainly based on the sequence-to-sequence architecture with specific designs for the characteristics of this task.
For example, \citet{ontology} employed clinical terms within the \textsc{Findings} as key information to enhance AIG.
Based on this work, \newcite{attend} further proposed to identify the importance of these clinical terms and selected the most salient ones to facilitate the recognition of significant content so as to improve the performance.
Although these efforts are able to find the important words to promote AIG, less attention is paid to the important aspect on leveraging the relation information among them.
For example, in Figure \ref{fig:example}, the observation word ``\textit{effusion}" and its modifier word ``\textit{moderate}" have a relation (which describes the severity of the symptoms) in between them, where such relation needs to appear in the \textsc{Impression}.
Therefore, to better generate \textsc{Impression}, in addition to using important words, it is required to recognize the relations among such words in the \textsc{Findings} and describe their corresponding relations for AIG.

In this paper, we propose to enhance AIG via a summarization model integrated with a word graph by leveraging salient words and their relations in the \textsc{Findings}.
In detail, the word graph is constructed by identifying the important words in the \textsc{Findings} and building connections among them via different typed relations.
To exploit the word graph, a \underline{W}ord \underline{G}raph guided \underline{Sum}marization model (\textbf{\textsc{WGSum}}) is designed to perform AIG, where the information from the word graph is integrated into the backbone decoder (e.g., LSTM \cite{see2017get} or Transformer \cite{liu2019text}) from two aspects, including enriching the decoder input as extra knowledge, as well as guiding the decoder to update its hidden states.
Experimental results illustrate that \textsc{WGSum} outperforms all baselines on two benchmark datasets, where the state-of-the-art performance is observed on all datasets in comparison with previous studies.
Further analyses also investigate how different types of edges in the graph affects the performance of our proposed model.

\section{The Proposed Method}

We follow the standard sequence-to-sequence paradigm for AIG.
In doing so, we regard the \textsc{Findings} as the source sequence $\mathbf{X} =  \left\{x_{1},...,x_{i},...,x_{N}\right\}$, where $N$ is the number of tokens in the \textsc{Findings}, and the goal of the task is to generate a target sequence (i.e., the \textsc{Impression}) $\mathbf{Y}=\left\{y_{1},...y_{i},..., y_{L}\right\}$, where $L$ is the number of tokens in the \textsc{Impression}.
An overview of our method is shown in Figure \ref{fig:architecture} where the details are illustrated in following subsections.

\subsection{The Overall Structure}
The model used in our method contains three major components, i.e., the \textsc{Findings} encoder, the graph encoder, and the graph guided decoder with their details and the training objective described below.

\vskip 0.5em
\noindent\textbf{\textsc{Findings} Encoder}~
Given a \textsc{Findings}, denoted by $\mathbf{X}$ with $N$ tokens, LSTM or the standard encoder from Transformer is applied to model the sequence and its output is the hidden state $\mathbf{h}^{x}$.
The process is formulated as
\begin{equation}
\setlength\abovedisplayskip{6pt}
\setlength\belowdisplayskip{6pt}
    \mathbf{h}^{x} = f_{fe}(x_1,...,x_i,...,x_N)
\end{equation}
where $f_{fe}(\cdot)$ refers to the \textsc{Findings} encoder.

\vskip 0.5em
\noindent\textbf{Graph Encoder}~
For the node $V$ and adjacency matrix $\mathbf{A}$ in the graph $G$ constructed from the \textsc{Findings}, we utilize graph neural networks\footnote{It can be any type (e.g., graph convolutional networks (GCN) and graph attention networks (GAT)).} (GNN) to encode them, because GNNs are powerful in encoding graph-like information \cite{chen-etal-2020-joint,zheng-kordjamshidi-2020-srlgrn,tian-etal-2021-aspect,tian-etal-2021-dependency}.
In detail, two encoders are employed to extract features from $G$, where one is used to construct the background information and the other is used to generate the dynamic guiding information.
The process is thus formalized as
\begin{align}
\setlength\abovedisplayskip{6pt}
\setlength\belowdisplayskip{6pt}
    \mathbf{z}^{b} &= f_{gb}(V,\mathbf{A}) \label{eq:zb}\\
    \mathbf{z}^{l} &= f_{gl}(V,\mathbf{A}) \label{eq:zl}
\end{align}
where $f_{gb}(\cdot)$ and $f_{gl}(\cdot)$ refer to two graph encoders, with $\mathbf{z}^{b}$ and $\mathbf{z}^{l}$ the intermeidate states used to generate the static background information and the dynamic guiding information, respectively.

\vskip 0.5em
\noindent\textbf{Graph Guided Decoder}~
In our model, $\mathbf{z}^{b}$ and $\mathbf{z}^{l}$ from the graph encoders are integrated into the backbone decoder (e.g., LSTM and Transformer decoder.) and perform a decoding process via
\begin{equation}
\setlength\abovedisplayskip{6pt}
\setlength\belowdisplayskip{6pt}
    P_{\text {vocab}} = f_{d}(\mathbf{h}^{x}, \mathbf{z}^{b},\mathbf{z}^{l})
    \label{eq:p_vocab}
\end{equation}
where $f_{d}(\cdot)$ represents the decoder.

\vskip 0.5em
\noindent\textbf{Objective}~
Since the \textsc{Findings} and the \textsc{Impression} are highly related, a pointer generator (PG) is also introduced to our model by

\begin{eqnarray}
\setlength\abovedisplayskip{6pt}
\setlength\belowdisplayskip{6pt}
    P(y_{t}\!\mid\!\mathbf{X}, y_{<t})\!\!=\!\!p_{\text {gen}}P_{\text {vocab}}\!\left(y_{t}\right)\!+\!({1\!-\!p_{\text{gen}}})\!\!\!\!\sum_{i:x_{i}\!=y_{t}}\!\!\!a_{i}^{t}
\end{eqnarray}
where $\mathbf{a}_{i}^{t}$ is the distribution over source tokens at step $t$, which is obtained by performing the attention mechanism on the source tokens; $p_{gen}$ and $(1-p_{gen})$ are the weights of predicting the next token from the vocabulary or the source tokens, respectively.
The model is then trained to maximize the negative conditional log-likelihood $P(\mathbf{Y} | \mathbf{X}, G)$ by of $\mathbf{Y}$ given $\mathbf{X}$ and $G$:
\begin{equation}
\setlength\abovedisplayskip{6pt}
\setlength\belowdisplayskip{6pt}
    \theta^{*}=\underset{\theta}{\arg \max}  \sum_{t=1}^{L} \log p\left(y_{t} \mid y_{1}, ...,  y_{t-1}, \mathbf{X}, G;\theta\right)
\end{equation}
where $\theta$ are the parameters of the model.

\subsection{Graph Construction}
In \textsc{Findings}, the most critical content that the radiologists need to summarize is the abnormality which usually includes the corresponding specific observations as well as their modifying words.
Therefore, in our study, we first extract 5 types of entities from \textsc{Findings}: \textit{anatomy}, \textit{observation}, \textit{anatomy modifier}, \textit{observation modifier}, and \textit{uncertainty}, where these entities are able to cover most of the key words that need to appear in \textsc{Impression}.
In addition, \newcite{attend} has shown that fine-grained words are more effective than the entire ontology. Inspiring by this idea, we regard words from these entities as nodes in our graph.
To avoid confusion, the repeated words are treated as one single node in each \textsc{Findings} even though they are not presented in the same entity.
Besides modifying relation, some other relations are also important for \textsc{Impression} generation, such as relations between \textit{anatomy} and \textit{observation}.
For example, in Figure \ref{fig:example}, the relation between ``\textit{pleural}" (\textit{anatomy}) and  ``\textit{effusion}" (\textit{observation}), which can be obtained for the detailed abnormality in the \textsc{Impression}.
To capture these types of relations, we leverage dependency trees which have been widely used to model word-word relations in many studies \cite{tian-etal-2020-improving-biomedical,tian-etal-2020-supertagging,pouran-ben-veyseh-etal-2020-improving-aspect,chen-etal-2021-relation}.
Thus, we define three types of edges for our word graph. Note that, each \textsc{Findings} has its corresponding word graph:
\vspace{-0.1mm}
\begin{itemize}[leftmargin=*]
\setlength{\topsep}{0pt}
\setlength{\itemsep}{0pt}
\setlength{\parsep}{0pt}
\setlength{\parskip}{0pt}
    \item \textbf{Type} \uppercase\expandafter{\romannumeral1}: this is the type using the natural order of words in an entity.
    In detail, we connect words if they are adjacent in the same entity.
    In Figure \ref{fig:example}, the pink dashed lines serve as the type I edge.
    For example, ``\textit{endotracheal tube}" is an entity, so that ``\textit{endotracheal}" is connected to ``\textit{tube}".
    \item \textbf{Type} \uppercase\expandafter{\romannumeral2}: this is the type using the relations among entities within the same category (e.g., observation and its modifier, and anatomy and its modifier).
    As shown in Figure \ref{fig:example}, given an \textit{observation},``\textit{effusion}" and a \textit{observation modifier} ``\textit{moderate}", the relation is constructed by connecting them with a
    green dash line.
   \item \textbf{Type} \uppercase\expandafter{\romannumeral3}: this is the type using the relations among entities across different categories (e.g., observation and anatomy).
    Different from the previous two types, this type is able to provide the global relation information while the previous two types emphasize the local information.
    In detail, we construct a dependency tree using stanza in the Universal Dependencies (UD) format \cite{nivre2020universal}.
   As shown in Figure \ref{fig:example}, given ``\textit{effusion}" and its head ``\textit{left}", they are connected with a purple dash line.
\end{itemize}
\noindent
Then the nodes and edges in graph are recorded by a node list $V$  and an adjacency matrix $\mathbf{A}$ which are then used as the input of graph encoder.
%
\subsection{Graph Guided Decoder}
We utilize graph to generate two different kinds of information and they are working on two aspects: enriching decoder input by static background information $\mathbf{h}^{b}$ and controlling decoder hidden state by dynamic guiding information $\mathbf{h}^{l}$, which are introduced in following parts.
\paragraph{Background Information}
Since the graph can be considered as a condensing of source sequence (i.e., \textsc{Findings}) which contains the most important information, it is appropriate to serve as a static background information to enrich the decoder inputs.
The first output $\mathbf{z}^{b}$ of graph encoder is used to construct the background information.
For the hidden state $\mathbf{z}_{i}^{b}$ in $\mathbf{z}^{b}$ of each node, we can obtain attention weights by:
\begin{align}
\setlength\abovedisplayskip{4pt}
\setlength\belowdisplayskip{4pt}
e_{i}^{b}&=\mathbf{p}_{b}^{\top} \tanh (\mathbf{W}_{b} \mathbf{z}_{i}^{b}+\mathbf{W}_{h} \mathbf{h}_{f}) \label{eq:attention2}\\
\mathbf{a}^{b}&=\text{softmax}(\mathbf{e}^{b})
\label{eq:attentiion3}
\end{align}
where $\mathbf{W}_{b}$,$\mathbf{W}_{h}$ and $\mathbf{p}_{b}$ are learnable parameters. 
For LSTM, we define $\mathbf{h}_{f}$ as the final hidden state, and for Transformer, we calculate the mean of all hidden states as $\mathbf{h}_{f}$. The attention distribution $\mathbf{a}^{b}$ can be viewed as a probability distribution over nodes in graph.
Next, $\mathbf{a}^{b}$ is used to produce a weighted sum of the nodes and then we obtain the static background information:
\begin{equation}
\setlength\abovedisplayskip{4pt}
\setlength\belowdisplayskip{4pt}
\mathbf{h}^{b}=\sum_{i} a_{i}^{b} \mathbf{z}_{i}^{b}
\label{background}
\end{equation}
For clarity, we simplify Equation (\ref{eq:attention2}), (\ref{eq:attentiion3}) and (\ref{background}) as a function $\text{AttCon}(\cdot)$. Therefore, $\mathbf{h}^{b}$ can be obtained by:
\begin{equation}
\setlength\abovedisplayskip{4pt}
\setlength\belowdisplayskip{4pt}
   \mathbf{h}^{b}= \text{AttCon}(\mathbf{z}^{b},\mathbf{h}_{f})
\end{equation}
In our model, the background information $ \mathbf{h}^{b}$ is directly concatenated to the decoder input.
For each decoder input $\mathbf{y}_{t-1}$ at step $t$, it is expanded as $\mathbf{y}_{t-1}^{\prime} = [\mathbf{y}_{t-1},\mathbf{h}^{b}]$.
%

\begin{table}[t]
\footnotesize
\centering
\setlength{\tabcolsep}{1.1mm}{\begin{tabular}{@{}l|rrr|rrr@{}}
\toprule
\multirow{2}{*}{\textsc{\textbf{Data}}} & \multicolumn{3}{c|}{\textsc{\textbf{OpenI}}}                                                                     & \multicolumn{3}{c}{\textsc{\textbf{MIMIC-CXR}}}                                                                    \\ \cmidrule(l){2-7} 
                                  & \multicolumn{1}{c}{\textsc{Train}} & \multicolumn{1}{c}{\textsc{Val}} & \multicolumn{1}{c|}{\textsc{Test}} & \multicolumn{1}{c}{\textsc{Train}} & \multicolumn{1}{c}{\textsc{Val}} & \multicolumn{1}{c}{\textsc{Test~}} \\ \midrule
\textsc{Report \#}                & 2,400                              & 292                              & 576                                & 122,014                            & 957                            & 1,606~                             \\
\midrule
\textsc{Afl}                & 37.89                              & 37.77                            & 37.98                              & 55.78                              & 56.57                            & 70.67                             \\ 
\textsc{Afs}                & 5.75                              & 5.68                            & 5.77                              & 6.50                              & 6.51                           &7.28~                             \\ 
\textsc{Afe}                & 18                              & 17                            & 18                              & 27                              &28                          &36~                             \\ 
\midrule
\textsc{Ail}                & 10.43                              & 11.22                            & 10.61                              & 16.98                             & 17.18                            & 21.71~                             \\ 
\textsc{Ais}                & 2.86                              & 2.94                            & 2.82                              & 3.02                              &3.04                          &3.49~                             \\

\bottomrule
\end{tabular}}
\caption{The statistics of the two benchmark datasets (random split for \textsc{OpenI} and official split for \textsc{MIMIC-CXR}) in the training, validation, and test sets, where the number of reports, the averaged sentence-based (\textsc{Afs}, \textsc{Ais}) and word-based length (\textsc{Afl}, \textsc{Ail}) of \textsc{Impression} and \textsc{Findings}, and the averaged number of edges in graph (\textsc{Afe}) are reported.}
\label{table:datasets}
\vskip -1em
\end{table}

\begin{table*}[t]
\footnotesize
\centering
\resizebox{.94\textwidth}{!}{
\begin{tabular}{l|p{3.5cm}|ccc|ccc}
\toprule[1pt]
\multirow{2}{*}{\textsc{\textbf{\makecell[c]{Data}}}} & \multirow{2}{*}{\textsc{\textbf{\makecell[c]{Model}}}} 
& \multicolumn{3}{c|}{\textsc{\textbf{ROUGE}}} & \multicolumn{3}{c}{\textsc{\textbf{FC}}}  \\  
& & \textsc{R-1}  & \textsc{R-2}  & \textsc{R-L}   &\textsc{\textbf{P}} & \textsc{\textbf{R}} &\textsc{\textbf{F-1}} \\
\midrule                       
\multirow{6}{*} {\makecell*[l]{\textsc{OpenI}}}
& \textsc{PG-LSTM} & {63.21} & {54.13} & {62.78} & {-} & {-} & {-}\\
& \textsc{WGSum} ({LSTM+GCN}) & {{63.69}} & {{54.88}}  & {{63.33}} & {-} & {-} & {-}\\
& \textsc{WGSum} ({LSTM+GAT}) & \textbf{{64.32}} & \textbf{{55.48}}  & \textbf{{63.97}} & {-} & {-} & {-}\\
\cmidrule(l){2-8}
& \textsc{PG-Trans} & {59.66} & {49.41} & {59.18} & {-} & {-} & {-}\\
& \textsc{WGSum} ({Trans+GCN}) & {60.95} & {50.67}  & {60.85} & {-} & {-} & {-}\\
& \textsc{WGSum} ({Trans+GAT}) & \textbf{{61.63}} & \textbf{{50.98}}  & \textbf{{61.73}} & {-} & {-} & {-}\\
\midrule
\multirow{6}{*} {\makecell*[l]{\textsc{MIMIC-CXR}}}
& \textsc{PG-LSTM} & {46.41} & {32.33} & {44.76} & {54.72} & {45.37} & {49.61}\\
& \textsc{WGSum} ({LSTM+GCN}) & {46.93} & {32.69}  & {45.25} & {55.23} & {46.21} & {50.32}\\
& \textsc{WGSum} ({LSTM+GAT}) & \textbf{{47.48}} & \textbf{{33.03}}  & \textbf{{45.43}} & \textbf{55.82} & \textbf{47.13}  & \textbf{51.11} \\
\cmidrule(l){2-8}
& \textsc{PG-Trans} & {47.16} & {32.31} & {45.47} & {56.18} & {49.08} & {52.39}\\
& \textsc{WGSum} ({Trans+GCN}) & {{47.93}} & {{32.63}}  & {{46.23}} & {56.37} & {50.84} & {53.46} \\
& \textsc{WGSum} ({Trans+GAT}) & \textbf{{48.37}} & \textbf{{33.34}}  & \textbf{{46.68}}  & \textbf{56.83} & \textbf{51.22}  & \textbf{53.88}\\
\bottomrule
 \end{tabular}
 }
\vskip -2mm
  \caption{The performance of the baselines and our proposed methods with different GNNs on \textsc{OpenI} and \textsc{MIMIC-CXR}. R-1, R-2 and R-L denote ROUGE-1, ROUGE-2 and ROUGE-L, respectively.}%
  \label{Tab:performance_on_different_base}
\vskip -4mm
\end{table*}

\paragraph{Dynamic Guiding Information}
Since $\mathbf{h}^{b}$ remains unchanged and works as the global static knowledge during the decoding process, to make the guidance more flexibly, the other information $\mathbf{z}^{l}$ from graph encoder in Equation (\ref{eq:zl}) is used to generate dynamic guiding information $\mathbf{h}_{t}^{l}$ for each decoding step t.
For different backbone decoders, there is a little difference in generating $\mathbf{h}_{t}^{l}$.
For LSTM decoder, each cell updates its information by two states and one input: cell state $\mathbf{c}_{t-1}$, hidden state $\mathbf{s}_{t-1}$ and input $\mathbf{y}_{t-1}^{\prime}$, which is formulated as:
\begin{equation}
\setlength\abovedisplayskip{4pt}
\setlength\belowdisplayskip{4pt}
    [\mathbf{c}_{t},\mathbf{s}_{t}^{\prime}] = \text{LSTM}(\mathbf{c}_{t-1},\mathbf{s}_{t-1},\mathbf{y}_{t-1}^{\prime})
\end{equation}
\noindent
where $\mathbf{c}_{t}$ usually contains rich contextual information and it is appropriate to compute guiding information $\mathbf{h}_{t}^{l}$.\footnote{ We also try to use hidden state $\mathbf{s}_{t}^{\prime}$ from LSTM to calculate $\mathbf{h}_{t}^{l}$, but the performance is not better than that of $\mathbf{c}_{t}$.}
\begin{equation}
\setlength\abovedisplayskip{4pt}
\setlength\belowdisplayskip{4pt}
    \mathbf{h}_{t}^{l}= \text{AttCon}(\mathbf{z}^{l},\mathbf{c}_{t})\end{equation}
For Transformer, the general decoder only has one hidden state $\mathbf{s}_{t}^{\prime}$ which is the output of the last layer.
In this part, we regard the output of the penultimate layer as another hidden state $\mathbf{c}_{t}$ which is then used to generate dynamic information $\mathbf{h}_{t}^{l}$ by:
\begin{align}
\setlength\abovedisplayskip{4pt}
\setlength\belowdisplayskip{4pt}
    \mathbf{h}^{l}_{t} = \text{softmax}(\mathbf{c}_{t}(\mathbf{z}^{l})^{\top}) \mathbf{z}^{l}
\end{align}
After obtaining the dynamic guidance $\mathbf{h}_{t}^{l}$ from the LSTM decoder or Transformer decoder, it is then utilized to update decoder hidden state $\mathbf{s}_{t}^{\prime}$ by:
\begin{equation}
\setlength\abovedisplayskip{4pt}
\setlength\belowdisplayskip{4pt}
    \mathbf{s}_{t} = \mathbf{s}_{t}^{\prime} + f_{g}(\mathbf{z}^{l}) \cdot f_{u}(\mathbf{z}^{l})\end{equation}
where $f_{g}(\cdot)$ and $f_{u}(\cdot)$ are fully connected layers.
\paragraph{Vocabulary Distribution}
To incorporate the \textsc{Findings} information for final prediction, we calculate the attention context vector $\mathbf{g}_{t}^{c}$ by the same way as Equation (\ref{eq:attention2}), (\ref{eq:attentiion3}) and (\ref{background}) using sequence encoder hidden state $\mathbf{h}^{x}$ as well as the updated $\mathbf{s}_{t}$:
\begin{equation}
\setlength\abovedisplayskip{4pt}
\setlength\belowdisplayskip{4pt}
    \mathbf{g}_{t}^{c}= \text{AttCon}(\mathbf{h}^{x},\mathbf{s}_{t}) \label{context vector}
\end{equation}
Then both $\mathbf{g}_{t}^{c}$ and decoder hidden state $\mathbf{s}_{t}$ are used to calculate vocabulary distribution at step $t$:
\begin{equation}
\setlength\abovedisplayskip{4pt}
\setlength\belowdisplayskip{4pt}
P_{vocab}=\text{softmax}\left(\mathbf{Q}^{\prime} \tanh \left(\mathbf{Q}\left[\mathbf{s}_{t} ; \mathbf{g}_{t}^{c}\right]\right)\right)
\label{eq:p_vocab_detail}
\end{equation}
where $\mathbf{Q}^{\prime}$ and $\mathbf{Q}$ are learnable weights.
\section{Experiment Settings}

\subsection{Dataset}
We conduct our experiments on the following two datasets of radiology reports: \textbf{\textsc{OpenI}} \cite{demner2016preparing} and \textbf{\textsc{MIMIC-CXR}} \cite{johnson2019mimic}.
The former is collected by Indiana University with 3,268 reports after pre-processing.
The latter contains a larger amount of data, where we obtain 124,577 after pre-processing.
In our experiments, the original \textsc{Impression}s written by the radiologists are considered as the ground truth.
For both datasets, we follow \newcite{zhang2018learning} and \newcite{attend} to exclude the following types of reports: (a) incomplete reports; (b) reports whose \textsc{Findings} have less than 10 words; and (c) reports with \textsc{Impression} words less than 2.
For the dataset splits, \textsc{OpenI} is partitioned into train/validation/test set by 2400:292:576 randomly in our experiments.
For \textsc{MIMIC-CXR}, we apply two splits, one is the official split as the dataset published by \newcite{johnson2019mimic}, and the other is a random split with a ratio of 8:1:1, which is the same as \newcite{attend}.
The statistics of the datasets are shown in Table \ref{table:datasets}.
%
%

\begin{table*}[t]
\begin{tabular}{l|rrr|rrr|rrr}
\toprule
{\multirow{3}{*}{\diagbox[width=10.4em,trim=l]
{ \textsc{\textbf{\makecell[c]{\\Model}}}}{\textsc{\textbf{\makecell[c]{Data\\}}}}}}  & \multicolumn{3}{c|}{\textsc{\textbf{OpenI}}} & \multicolumn{6}{c}{\textsc{\textbf{MIMIC-CXR}}} \\
\cmidrule(r){2-10}
& \multicolumn{3}{c|}{\textsc{\textbf{Random split}}} & \multicolumn{3}{c|}{\textsc{\textbf{Official split}}}   & \multicolumn{3}{c}{\textsc{\textbf{Random split}}} \\  
                                                                                   & \textsc{R-1}  & \textsc{R-2}  & \textsc{R-L}   & \textsc{R-1}        & \textsc{R-2}       & \multicolumn{1}{c|}{\textsc{R-L}}         & \textsc{R-1}        & \textsc{R-2}       & \textsc{R-L}      \\
\cmidrule(lr){1-10}
\textsc{LexRank}   &14.63    &4.42    &14.06    &18.11    & 7.47    & 16.87      & -      & -       & -  \\

\textsc{TransformerExt}& 15.58    & 5.28     & 14.42    & 31.00   &16.55   &27.49   & -   & -   & -      \\

\textsc{CAVC}   & 53.18   & 39.59    & 52.86    &43.97    & 29.36     & 42.50    & -    & -     & -     \\
\textsc{CGU}   & 61.60   & 53.00    & 61.58    &46.50    & 32.61     & 44.98    & -    & -     & -     \\
$\textsc{OntologyABS}^{\dagger}$ & -  & -   & -   & -    & -    & -  & 53.57   &40.78  & 51.81 \\
\cmidrule(lr){1-10}
\textsc{WGSum (LSTM+GAT)} &\textbf{{64.32}} & \textbf{{55.48}}  & \textbf{{63.97}} &{47.48} &{33.03} &{45.43} & {54.97}  & {43.64}  & {53.81} \\

\textsc{WGSum (Trans+GAT)}&{61.63} & {50.98}  & {61.73}   & \textbf{{48.37}} & \textbf{{33.34}}  & \textbf{{46.68}}       & \textbf{56.38}    & \textbf{44.75}   & \textbf{55.32}  \\

\bottomrule
\end{tabular}
 \vspace{-2mm}
\caption{Comparisons of our proposed models with previous study on the test sets of \textsc{IU X-Ray} and \textsc{MIMIC-CXR} with respect to ROUGE metric. $\dagger$ refers to that the results is directly cited from the original paper.}
\label{Tab:performance_on_different_model}
\vskip -1em
\end{table*}

\subsection{Baseline and Evaluation Metrics}
To compare the performance of our proposed models, we use the following models as our baselines:
\vskip -1em
\begin{itemize}[leftmargin=*]
    \setlength{\topsep}{0pt}
    \setlength{\itemsep}{0pt}
    \setlength{\parsep}{0pt}
    \setlength{\parskip}{0pt}
    \item \textbf{\textsc{PG-LSTM}} \cite{see2017get}: This is Pointer Generator Network (PGN) with copy mechanism where both encoder and  decoder are vanilla LSTMs without graph information.
    \item \textbf{\textsc{PG-Trans}} \cite{liu2019text}: This is a Transformer-based model where both the encoder and decoder are replaced with the transformer and the copy mechanism is removed.
\end{itemize}
\noindent
Besides, we also compare our model with those in previous studies, including extractive summarization models, e.g., \textbf{\textsc{LexRank}} \cite{erkan2004lexrank}, \textbf{\textsc{TransformerEXT}} \cite{liu2019text}, as well as abstractive summarization models, e.g., \textbf{\textsc{CAVC}} \cite{song2020controlling}, \textbf{\textsc{CGU}} \cite{lin2018global} and \textbf{\textsc{OntologyABS}} \cite{attend}.
In our experiments, we use ROUGE metrics \cite{lin2004rouge} to evaluate the generated \textsc{Impression}s. We only report $\textsc{F}_{1}$ scores of ROUGE-1 (R-1), ROUGE-2 (R-2) and ROUGE-L (R-L), where R-1, R-2 are unigram and bigram overlap measuring the informativeness and R-L is the longest common sub-sequence overlap aiming to assess fluency.
In addition, to evaluate the factual consistency (FC), CheXbert \cite{smit2020chexbert}\footnote{FC only apply to MIMIC-CXR since the CheXbert is designed for MIMIC-CXR and we obtain Chexbert from \url{https://github.com/stanfordmlgroup/CheXbert}} is utilized to detect 14 observations related to diseases in reference impressions and generated impressions. Then precision, recall and F1 score are used to evaluate the performance.
\subsection{Implementation Details}
We employ stanza \cite{zhang2020biomedical}\footnote{Stanza provides packages to process the clinical text: \url{https://stanfordnlp.github.io/stanza/}.}, a python-based natural language processing library, to recognize named entities and get the syntactic analysis. Then we use the extracted entities and dependency tree to construct graph for each \textsc{Findings}.
We implement our model based on \newcite{zhang2018learning}\footnote{\url{https://github.com/yuhaozhang/summarize-radiology-findings}} and \newcite{liu2019text}\footnote{\url{https://github.com/nlpyang/PreSumm}}.
Since the quality of text representation plays an important role in model performance \cite{mikolov2013efficient,song-etal-2017-learning,song-etal-2018-directional,peters-etal-2018-deep,ijcai2018-607,devlin2019bert,joshi-etal-2020-spanbert,song2021zen}, we try two powerful \textsc{Findings} encoders, namely, LSTM and Transformer, which have achieved state-of-the-art results in many natural language processing tasks.
For \textsc{WGSum} (LSTM+GAT)), we employ 2-layer GAT with hidden size of 200 as our graph encoder, 2-layer Bi-LSTM encoder for findings sequence with hidden size of 100 for each direction and 1-layer LSTM for decoder with hidden size of 200.
The dropout is set to 0.5 for embedding layer.
We use Adam optimizer \cite{kingma2014adam} with the learning rate of 0.001.
For \textsc{WGSum(Trans+GAT)}, the graph encoder is a 2-layer GAT with hidden size 512 and the \textsc{Findings} encoder is a 6-layer Transformer with 512 hidden size and 2,048 feed-forward filter size. The decoder is also a 6-layer Transformer with hidden size 512.

\begin{figure}[t]
\centering
\includegraphics[width=0.5\textwidth, trim=0 25 0 0]{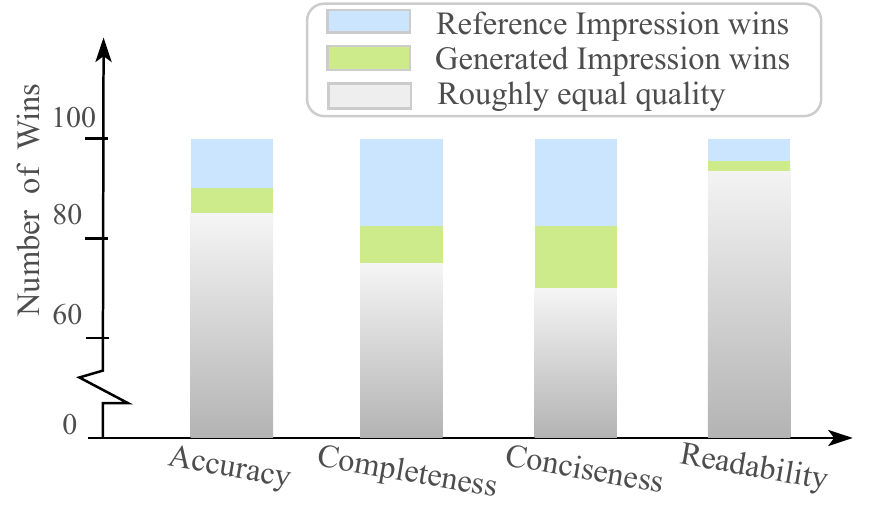}
\vspace{0.1em}
\caption{Experts evaluation results of 100 sampled test examples. For each of the metrics, more than 80\% of the generated \textsc{Impression}s are at least as good as the reference \textsc{Impression}s.}
\label{fig:human_evaluation}
\vskip -1em
\end{figure}

\section{Results and Analyses}

\subsection{Effect of word graph}
To illustrate the validity of the word graph, we conduct experiments with the aforementioned baselines on the two benchmark datasets.
Besides, we also try two different types of GNNs: GCN \cite{kipf2016semi} and GAT \cite{velivckovic2017graph} respectively.
The results are shown in Table \ref{Tab:performance_on_different_base} and there are several observations drawn from different aspects.
First, we can observe that for our word graph, the encoder GAT is more effective than GCN where GAT can bring more significant improvements on the two datasets.
The main reason might be that GAT is more powerful in updating node representation via self-attention.
Second, integrating the word graph into the two different PGNs gains better performance on both the datasets, which confirms the usefulness of the word graph.
Third, for \textsc{OpenI}, the LSTM-based models outperform much more than the Transformer-based models, while on \textsc{MIMIC-CXR}, the Transformer-based models are more effective. The main reason could be that the LSTM is able to obtain prominent performance in the small dataset and the Transformer is more powerful under a large amount of data.
Fourth, on the FC metrics on \textsc{MIMIC-CXR}, our proposed methods also outperform the \textsc{Base} model, indicating that the generated \textsc{Impression}s from our methods are more accurate and reasonable, which is because the word graph can provide both key word and relation information for the generation process so that the decoder tends to produce words with correct relations.
%
%

\subsection{Comparison with Previous Studies}
In this subsection, we compare our models with existing studies on the two datasets and report all results (i.e., ROUGE scores) in Table \textcolor{purple}{\ref{Tab:performance_on_different_model}}. 
We can get several observations.
First, the abstractive models are apparently more effective than the extractive models in AIG, owing to the characteristics of \textsc{Findings} and \textsc{Impression} in radiology reports.
Second, our models with the word graph show the effectiveness of both key words information and their relations in this task when being compared to the previous models that only leverage medical term information, e.g., \textsc{OntologyABS} only uses ontology information in database RadLex\footnote{\url{http://www.radlex.org/Files/rad lex3.10.xlsx}}.
Third, our methods achieve the best performance among all previous models, which demonstrates that using background knowledge and dynamic guidance information to control the decoding process is an appropriate design to improve the quality of the generated \textsc{Impressions}.
%
%
\begin{figure}[t]
\centering
\includegraphics[width=0.5\textwidth, trim=0 25 0 0]{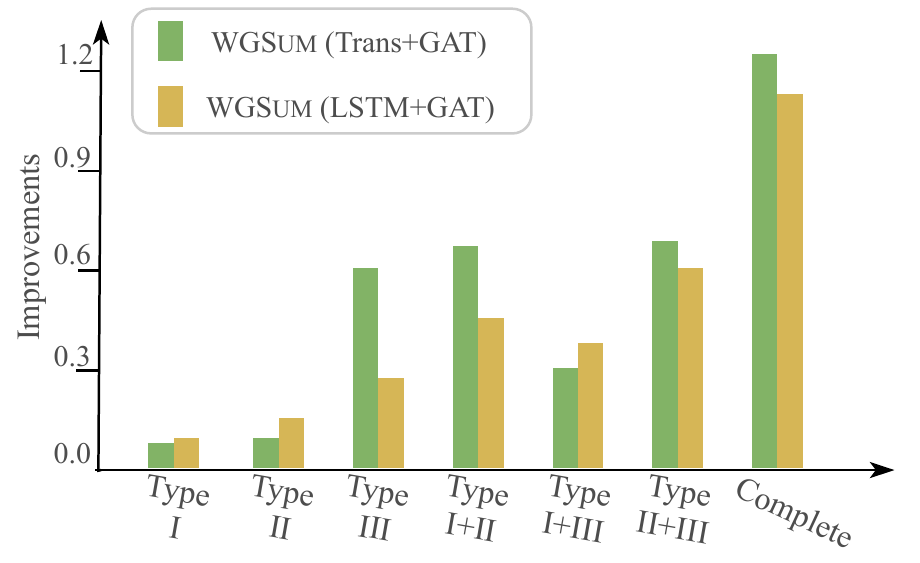}
\vspace{0.1em}
\caption{The improvements (R-1) of \textsc{WGSum} with different graph edges on MIMIC-CXR dataset.}
\label{fig:edges_types}
\vskip -1em
\end{figure}

\subsection{Expert Evaluation}
Since the limitation of ROUGE metrics, we further conduct an expert evaluation for a better understanding of the generated \textsc{Impression}s.
We randomly select 100 generated \textsc{Impression}s along with their corresponding reference \textsc{Impression}s and \textsc{Findings} from \textsc{MIMIC-CXR}.
To avoid potential bias, we randomly order the predicted and reference \textsc{Impression}s. We extend \newcite{attend} metrics to four metrics: Accuracy, Completeness, Conciseness and Readability.
Three medical experts are employed to score each \textsc{Impression} on these metrics.
Figure \ref{fig:human_evaluation} presents the results of human evaluation. We can observe that although reference \textsc{Impression}s written by the radiologists are still better, there are still over 80\% of generated \textsc{Impression}s have roughly equal or better quality.
About 85\%, 75\%, 70\%, and 94\% of generated \textsc{Impression}s are equal to human written \textsc{Impression}s on the four metrics (Accuracy, Completeness, Conciseness, and Readability).
In addition, there are even 5\%, 7\%, 12\%, 2\% of generated \textsc{Impression}s surpassed the reference \textsc{Impression}s on these metrics.

\begin{figure}[tb]
\centering
\includegraphics[width=0.47\textwidth, trim=0 30 0 0]{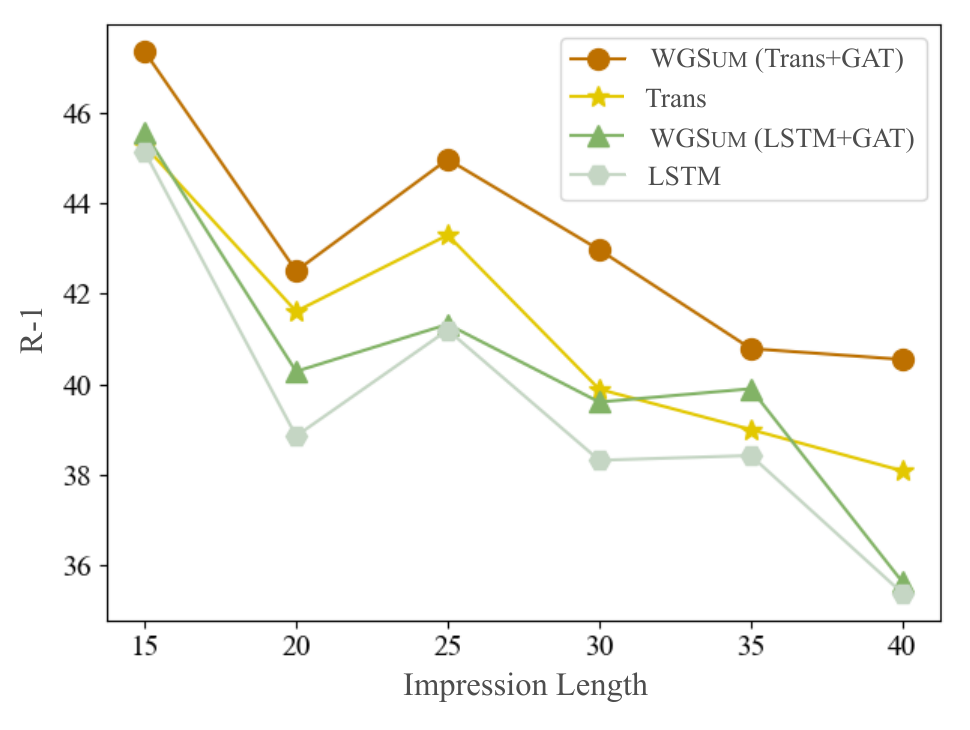}
\vspace{1em}
\caption{The performance (R-1) of generated \textsc{impression}s from \textsc{WGSum (Trans+GAT)}, \textsc{PG-Trans}, \textsc{WGSum (LSTM+GAT)} and \textsc{PG-LSTM} on the \textsc{\textsc{MIMIC-CXR}} test set. Each group contains \textsc{Impression}s of different length intervals.}
\label{fig:impression_length}
\vskip -1.em
\end{figure}

\begin{figure*}[t]
\centering
\includegraphics[width=1\textwidth, trim=0 30 0 0]{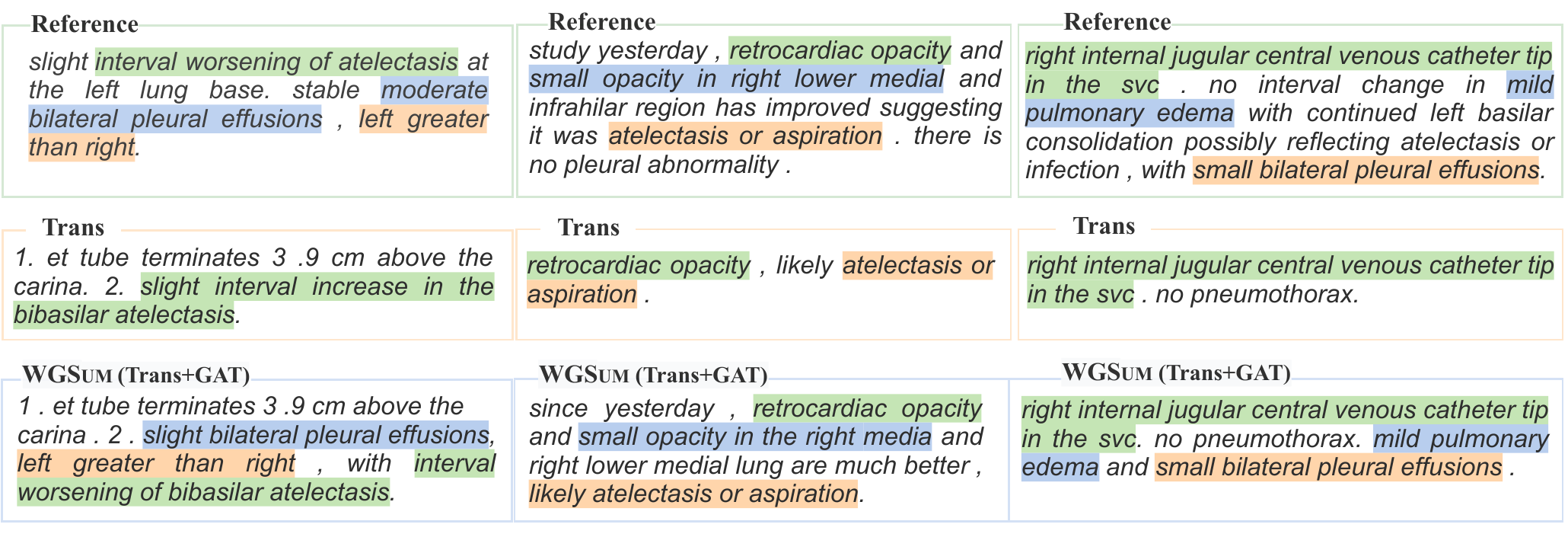}
\vspace{0.1em}
\caption{Examples of the generated \textsc{Impression}s from two models (i.e., the \textsc{PG-Trans} and the \textsc{WGSum (Trans+GAT)}, respectively) , as well as the reference \textsc{Impression}s.}
\label{fig:case_study}
\vskip -1em
\end{figure*}

\subsection{Analyses}
We conduct further analyses on Graph Edge, \textsc{Impression} Length, and Case Study.

\paragraph{Graph Edge}
As we introduced before, our graph contains three types of edges, i.e., entity interval edge (Type \uppercase\expandafter{\romannumeral1}), entity modifier edge (Type \uppercase\expandafter{\romannumeral2}), and edge from dependency tree (Type \uppercase\expandafter{\romannumeral3}). To show the effect of different edges, we conduct experiments for \textsc{WGSum} (LSTM+GAT) and \textsc{WGSum} (Trans+GAT) with different edges on \textsc{MIMIC-CXR}. The improvements from these different edge combinations are shown in Figure \ref{fig:edges_types}.
First, we can observe that models incorporating word graph outperform the baselines no matter with what type of edge, indicating the effectiveness of our innovation in combining the entity words and their relations into the word graph.
Second, regardless of \textsc{WGSum} (LSTM+GAT) or \textsc{WGSum} (Trans+GAT), Type \uppercase\expandafter{\romannumeral3} edge can bring the most significant improvements, while Type \uppercase\expandafter{\romannumeral1} brings little improvements. The main reason might be that the dependency tree contains more comprehensive and accurate relations for the entity words from \textsc{Findings}.
Third, \textsc{WGSum} (Trans+GAT) usually obtains better improvements than \textsc{WGSum} (LSTM+GAT) in most cases.
\paragraph{\textsc{Impression} Length} 


Another factor that could affect the model performance is the number of tokens in the \textsc{Impression}.
To test the effect of \textsc{Impression} length, we categorize all generated \textsc{Impression}s in the \textsc{MIMIC-CXR} test set into 6 groups (within [15,40] with the interval of 5) and compare the R-1 score for each group. The results are shown in Figure \ref{fig:impression_length}.
There are several observations.
First, when \textsc{Impression} gradually increases in the length, the performance of all models shows a downward trend, which indicates that long \textsc{Impression} generation is difficult for all models.
Second, \textsc{Transformer}-based methods are more effective than LSTM based models, especially for long \textsc{Impression}.
The main reason might be that the Transformer is more powerful in dealing with longer sequences via its self-attention mechanism.
Third, both \textsc{WGSum (Trans+GAT)} and \textsc{WGSum (LSTM+GAT)} show their superiority when being compared to their baselines and obtain better results in almost all of the groups.

\paragraph{Case Study}
To further analyze the effect of our proposed model, we perform qualitative analysis on some cases with their reference and generated \textsc{Impression}s from different models.
Figure \ref{fig:case_study} shows three cases from \textsc{MIMIC-CXR} where different colors on text refer to varied key information.
It is found that in the first case, when referring to the corresponding \textsc{Findings} our model generates more complete \textsc{Impression} than the reference, e.g., \textit{``et tube terminates 3 .9 cm above the carina''} is a helpful text piece but does not appear in reference \textsc{Impression}.
In addition, compared to the reference \textsc{Impression}s written by radiologists, our method covers almost all of the key information in the generated \textsc{Impression}s.
For example, the key information \textit{``moderate bilateral pleural effusions"},\textit{"mild pulmonary edema"} and \textit{``small opacity in the right media "} in the three examples are not generated in \textsc{PG-Trans} model, but they are necessary for describing the clinical condition.
\section{Related Work}
%
Our work focuses on summarizing the \textsc{Findings} of radiology reports to generate the \textsc{Impression}, which is essentially an abstractive summarization task.
For abstractive summarization, there exists a serious problem known as hallucination \cite{maynez2020faithfulness}, in which the generated summary contains fictional content. This problem also exists in AIG and would lead to misdiagnosis for the patient.
To tackle this problem, in the general domain, many attempts have been made in terms of guiding information to control the generation process and output the high-quality summary \cite{li2018guiding, hsu2018unified, pilault2020extractive, huang2020knowledge, haonan2020exploring}.
For the \textsc{Impression} generation task in the medical domain, there also exist several solutions.
\newcite{zhang2018learning} encodes a section of the radiology report as the background information to guide the decoding process.
\newcite{ontology} employs the entire ontological terms extracted from \textsc{Findings} as the medical terms, and then enhances the summarizer by selecting the salient information.
\newcite{attend} further splits ontological terms into words and then incorporates these words into summarization by a separate encoder.
Compared to these studies, our model offers an alternative solution to robustly enhancing guidance with a word graph for summarizing the \textsc{Findings} of radiology reports without requiring external resources.
To our best knowledge, this is the first work employing word relation information for AIG.

\section{Conclusion}
In this paper, we propose a novel method for AIG, where a word graph is constructed from the \textsc{Findings} by identifying the salient words and their relations and a graph-based model \textsc{WGSum} is designed to generate \textsc{Impression}s with the help of the word graph.
In doing so, the information from the word graph guides the decoding process with the help of background information and dynamic guiding information.
Experimental results on two benchmark datasets show the validity of our proposed method, which obtains the state-of-the-art performance on both datasets.
Further analyses on the effect of edge types demonstrate that our model can generate \textsc{Impression} with accurate medical items.

\section*{Acknowledgements}
This work is supported by Chinese Key-Area Research and Development Program of Guangdong Province (2020B0101350001) and also partially supported by NSFC under the project ``The Essential Algorithms and Technologies for Standardized Analytics of Clinical Texts'' (12026610). 
We thank three reviewers from Sun Yat-sen University: Mei Feng, Xiang Chuqi, and Liang Feiyan.

\bibliographystyle{acl_natbib}
\bibliography{acl2021}

\end{document}